\documentclass{article}

\usepackage{PRIMEarxiv}

\usepackage[utf8]{inputenc} 
\usepackage[T1]{fontenc}    
\usepackage{hyperref}       
\usepackage{url}            
\usepackage{booktabs}       
\usepackage{amsfonts}       
\usepackage{nicefrac}       
\usepackage{microtype}      
\usepackage{lipsum}
\usepackage{fancyhdr}       
\usepackage{graphicx}       
\graphicspath{{media/}}     
\usepackage{booktabs}
\usepackage{multirow}
\title{ERNIE-ViL 2.0: Multi-view Contrastive Learning for Image-Text Pre-training}

\author{
     Bin Shan
\hspace{1cm} Weichong Yin
\hspace{1cm} Yu Sun 
\hspace{1cm} Hao Tian  
\hspace{1cm} Hua Wu
\hspace{1cm} Haifeng Wang
\\\\\texttt{Baidu Inc., China}
\\\\
\texttt{\{shanbin01, yinweichong, sunyu02\}@baidu.com} 
}

\begin{document}
\maketitle
\begin{abstract}
Recent Vision-Language Pre-trained (VLP) models based on dual encoder have attracted extensive attention from academia and industry due to their superior performance on various cross-modal tasks and high computational efficiency. They attempt to learn cross-modal representation using contrastive learning on image-text pairs, however, the built inter-modal correlations only rely on a single view for each modality.
Actually, an image or a text contains various potential views, just as humans could capture a real-world scene via diverse descriptions or photos. In this paper, we propose ERNIE-ViL 2.0, a Multi-View Contrastive learning framework to build intra-modal and inter-modal correlations between diverse views simultaneously, aiming at learning a more robust cross-modal representation.
Specifically, we construct multiple views within each modality to learn the intra-modal correlation for enhancing the single-modal representation. Besides the inherent visual/textual views, we construct sequences of object tags as a special textual view to narrow the cross-modal semantic gap on noisy image-text pairs. 
Pre-trained with 29M publicly available datasets, ERNIE-ViL 2.0 achieves competitive results on English cross-modal retrieval. Additionally, to generalize our method to Chinese cross-modal tasks, we train ERNIE-ViL 2.0 through scaling up the pre-training datasets to 1.5B Chinese image-text pairs, resulting in significant improvements compared to previous SOTA results on Chinese cross-modal retrieval. We release our pre-trained models in \url{https://github.com/PaddlePaddle/ERNIE/}.
\end{abstract}

\section{Introduction}
\label{sec: Introduction}

In the past two years, the Vision-Language Pre-training (VLP) models have achieved remarkable improvements on various cross-modal tasks such as Visual Question Answering (VQA) and cross-modal retrieval.
Most existing works \cite{li2020unicoder,chen2020uniter,Li2020OscarOA}, based on cross-modal transformer encoders, focus on designing multiple proxy pre-training tasks (e.g., Masked Language Modeling (MLM) and Masked Region Modeling (MRM)) to learn joint cross-modal representation. However, cross-modal attention layers in the encoder aim to fuse multiple token-level visual/textual features to learn the joint representation with massive interactions, resulting in high computation costs for real-life systems such as the online cross-modal retrieval system.
To address the limitation, recent works \cite{Radford2021LearningTV,Jia2021ScalingUV,DBLP:journals/corr/abs-2111-07783} based on dual-encoder architecture utilize a compute-efficient framework with light cross-modal interaction, achieving comparable results on vision-language tasks via training on large-scale image-text pairs.

However, they attempt to build the cross-modal alignment via single-view contrastive learning since the built inter-modal correlation only relies on a single view for each modality. Indeed, the intra-modal correlation which they neglect could enhance the single-modal representation and contribute to building a better cross-modal alignment. Besides, there often exists weak correlations in the noisy web-crawled image-text pairs containing inherent visual/textual views, widening the cross-modal semantic gap.

\begin{figure}
\centering\includegraphics [scale=0.6]{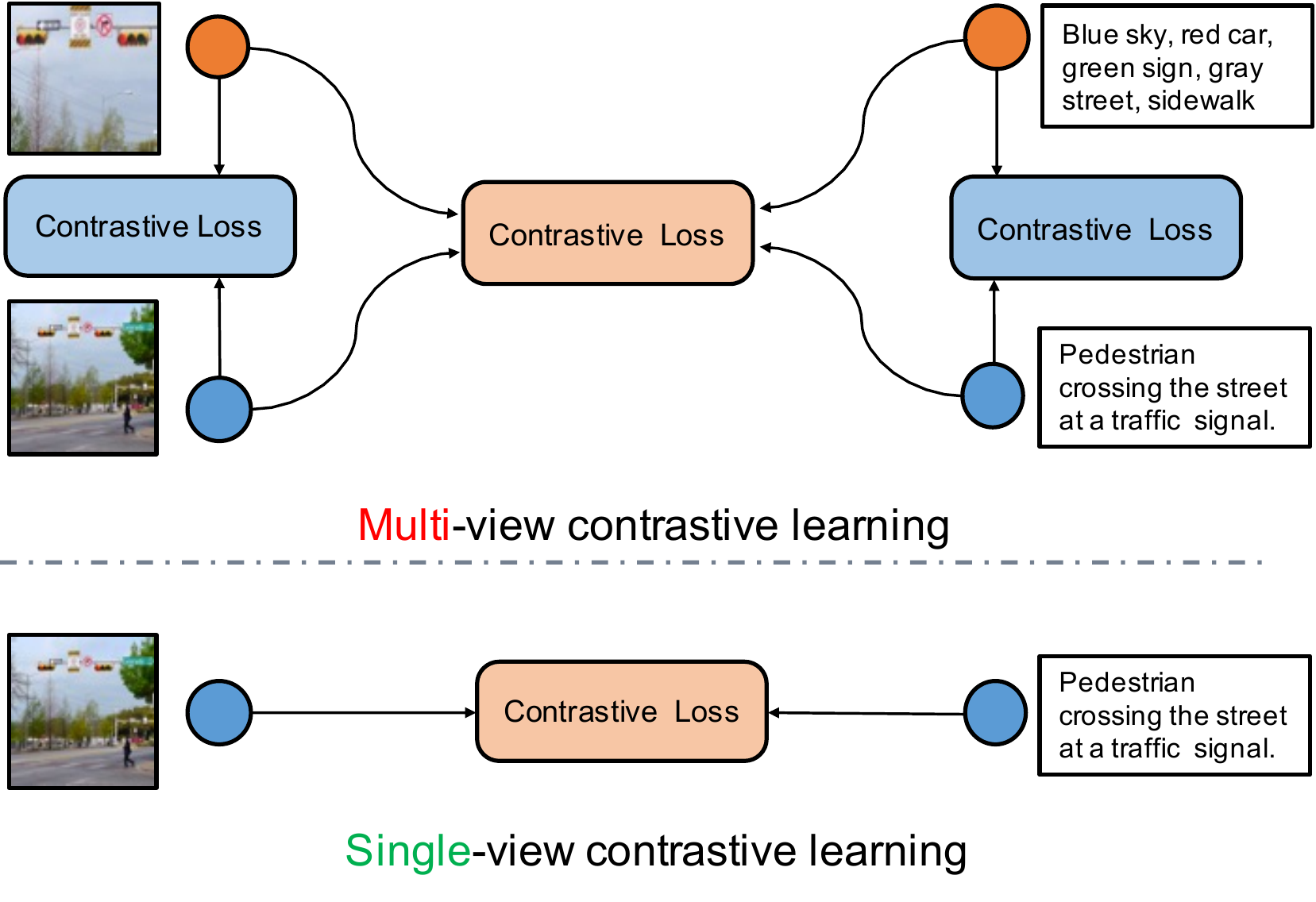}
\caption{Multi-view contrastive learning vs. Single-view contrastive learning. Single-view contrastive learning only focuses on a single inter-modal correlation between a visual and a textual view. Multi-view contrastive learning could learn on various kinds of intra-modal and inter-modal correlations through constructing diverse potential views. }
\label{fig:figure1_cap}
\end{figure}

In this paper, we propose ERNIE-ViL 2.0, a multi-view contrastive learning framework for cross-modal retrieval, aiming to learn robust cross-modal representation through modeling both inter-modal and intra-modal correlations between diverse views. As illustrated in Figure \ref{fig:figure1_cap}, compared to the conventional single-view contrastive learning methods, multi-view contrastive learning learns on both intra-modal and inter-modal correlations between various views, improving the robustness and generalization of the model. Similarly, CMC \cite{Tian2020ContrastiveMC} utilizes multi-view contrastive learning for visual representation learning, resulting in a more robust representation. Our method constructs various visual/textual views to enhance the representations within and across modalities. Specifically, for intra-modal contrastive view pairs, we construct image-image pairs and text-text pairs to enhance representation with each modality. In addition to the inherent visual/textual views, we construct sequences of object tags as a special textual view to mitigate the effects of noisy multi-modal data and ease the learning of vision-language alignment.

Based on the dual-encoder architecture, we train an English model on 29M publicly available datasets and achieve competitive results on cross-modal retrieval tasks. We further scale up the training datasets to 1.5B Chinese image-text pairs, resulting in significant improvements compared to previous SOTA results on Chinese cross-modal retrieval.

Overall, we summarise our contributions as three folds:
\begin{enumerate}
\item We propose the first multi-view learning framework for cross-modal retrieval, which exploits diverse views to obtain view-invariant and robust cross-modal representations.
\item We introduce object tags as special textual views, which effectively narrow the semantics gap between image-text and ease the learning of cross-modal alignment on large-scale noisy data. 
\item We establish a solid and comparable benchmark for English cross-modal retrieval using only noisy publicly available datasets. Furthermore, trained on 1.5B Chinese image-text pairs, our model achieves SOTA results on Chinese cross-modal retrieval.
\end{enumerate}

\section{Related Work}
\label{sec: Relate}
\subsection{Vision-Language Pre-training}
To improve the performance of downstream multi-modal tasks, Vision-Language Pre-training (VLP) models aim at aligning the representations in different modalities to the common space. According to the interaction between vision-language, they could be summarized into two categories: cross-encoder models and dual-encoder models.

Cross-encoder models \cite{li2020unicoder,chen2020uniter,Yu2021ERNIEViLKE,Li2020OscarOA,Li2021AlignBF,sun-etal-2021-lightningdot} build cross-modal attention between vision and language to achieve a joint multi-modal representation. They usually concatenate region-level visual features and token-level textual features as inputs of the transformer-based encoder and learn cross-modal joint representation using multiple proxy loss (e.g., MLM, MFM). These methods greatly improve the performance of various downstream tasks, especially for complexity reasoning V+L tasks (e.g., VQA). However, as the key ingredient, cross-modal attention brings huge computation costs in training and deployment. As mentioned in \cite{sun-etal-2021-lightningdot}, the inference time of UNITER \cite{chen2020uniter} with 12/24-layer is 48 seconds for text query from 5K COCO \cite{Chen2015MicrosoftCC} images.

Dual-encoder models \cite{Jia2021ScalingUV,Radford2021LearningTV,Yuan2021MultimodalCT,DBLP:journals/corr/abs-2111-07783,Jain2021MURALMM} employ a light cross-modal interaction, which only compute similarity  (e.g., cosine similarity) between visual and textual feature generated from separate encoder once, resulting in significantly lowering computational costs. Utilizing contrastive learning on large-scale web-crawl image-text pairs, they achieved several promising results on cross-modal tasks and performed remarkable zero-shot abilities. Besides, recent works \cite{singh2021flava,yuan2021florence,Li2021AlignBF,yu2022coca} attempt to combine dual-encoder and cross-encoder into an unified framework to learn cross-modal representation. However, they mainly rely on the dual-encoder modules when applied to cross-modal retrieval tasks.

Our method falls into the dual-encoder category and we proposes a multi-view contrastive learning framework to improve the performance of cross-modal tasks.
Recent works \cite{Yuan2021MultimodalCT,mu2021slip} are closely related to our work, which utilize a similar pre-training framework but focus on learning visual representation. Besides, we consider object detection tags as text views, while they use the tags as a supervisor to construct positive image samples.

\subsection {Multi-view Learning}
Multi-view learning exploits the consistency and complementary properties of different views to acquire the better generalization ability of models than single-view learning \cite{Xu2013ASO}. It is a general term comprising many methods: co-training \cite{Blum1998CombiningLA} and multi-kernel learning \cite{Cortes2009LearningNC}. 
Close to the self-supervised representation learning discussed in this paper, CMC \cite{Tian2020ContrastiveMC} and CPC \cite{hassani2020contrastive} attempt to learn more robust single-modal representation through using multi-view contrastive learning (e.g., future and past of sequential data in CPC, images in different color space in CMC). Whereas these methods present multiple definitions of views, the common insights of these works is learning view-invariant and more robust representations across various views through maximizing their co-occurrence semantics. 
While these works attempt to learn single-modal representation using multi-view learning, we focus on improving the robustness and generalization of cross-modal representation by simultaneously learning correlations between intra-modal and inter-modal views.

\section{Method}
\label{sec:Method}

In this section, we present ERNIE-ViL 2.0, a Multi-View Contrastive learning framework based on dual-encoder architecture for cross-modal retrieval. We first introduce the overview of the proposed framework in Section \ref{subsec:Architecture}, then describe how to construct various views in Section \ref{subsec:Construction}. Finally, the multi-view contrastive loss will be presented in Section \ref{subsec:Objectives}.

\subsection{Framework Overview}
\begin{figure*}
\centering\includegraphics [scale=0.55] {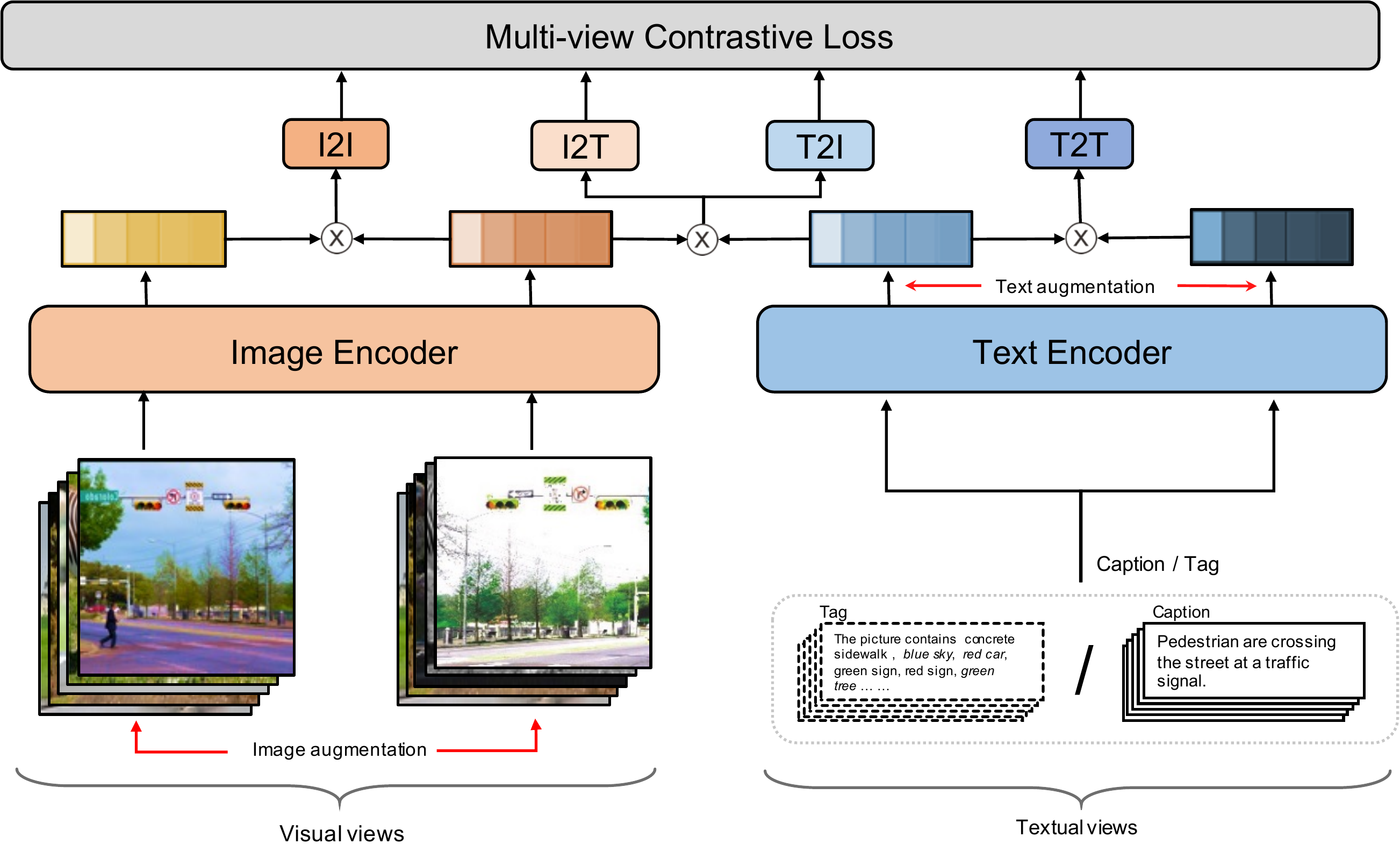}
\caption{The overview of our proposed Multi-View Contrastive learning framework (\textbf{ERNIE-ViL 2.0}). ERNIE-ViL 2.0 contains an image encoder and a text encoder. At each pre-training iteration, two visual views (images) and two textual views(captions or tags) are fed into the corresponding encoder. Particularly, we randomly feed one of the texts (captions or tags) into the encoder at each iteration. Multi-view contrastive loss sums four contrastive losses (I2I: image-image, I2T:image-text, T2I: text-image, T2T: text-text) calculated between different views with different weights.}
\label{fig:framework}
\end{figure*} 
\label{subsec:Architecture}
As illustrated in Figure 2, based on dual-encoder architecture, ERNIE-ViL 2.0 learns vision-language representation via modeling intra-modal and inter-modal correlations of diverse views simultaneously using multi-view contrastive learning. For learning the intra-modal correlations, we implement data augmentation to construct views for each modality. In visual modality, similar to SimCLR \cite{chen2020simple}, we utilize image augmentation (e.g. random cropping, jitter) to construct visual views. In textual modality, inspired by SimCSE \cite{gao2021simcse}, we consider dropout noise as text augmentation to construct textual views. For learning the inter-modal correlations, besides inherent image-caption pairs contained in datasets, we introduce the special textual sequence (object tags) to improve the inter-modal alignment.

We propose a multi-view contrastive loss to simultaneously learn the intra-modal and inter-modal alignments across multiple views, consisting of I2I (image-image), I2T (image-text), T2I (text-image), and T2T (text-text).
\subsection{Construction of Diverse Views}
\label{subsec:Construction}
Constructing effective and diverse views from image-caption pairs is the essential ingredient of our proposed framework. Overall, we construct six views for visual and textual modality, containing four regular views for the image and caption and two special views built from sequences of object tags. 

To construct the visual views, image augmentation (e.g., random crop and jitter) as a widely used technique for visual representation learning is utilized to generate different visual views. Thus, we implement random image augmentation on the same image twice to get two different visual views defined as $I_{v_1}$ and $I_{v_2}$. Since SimCSE\cite{gao2021simcse} uses dropout as a text augmentation and presents a promising and steady performance, we feed captions to a text encoder with dropout units twice to obtain different textual views $T_{v_1}$ and $T_{v_2}$. 

\begin{figure}
\centering\includegraphics [scale=0.35] {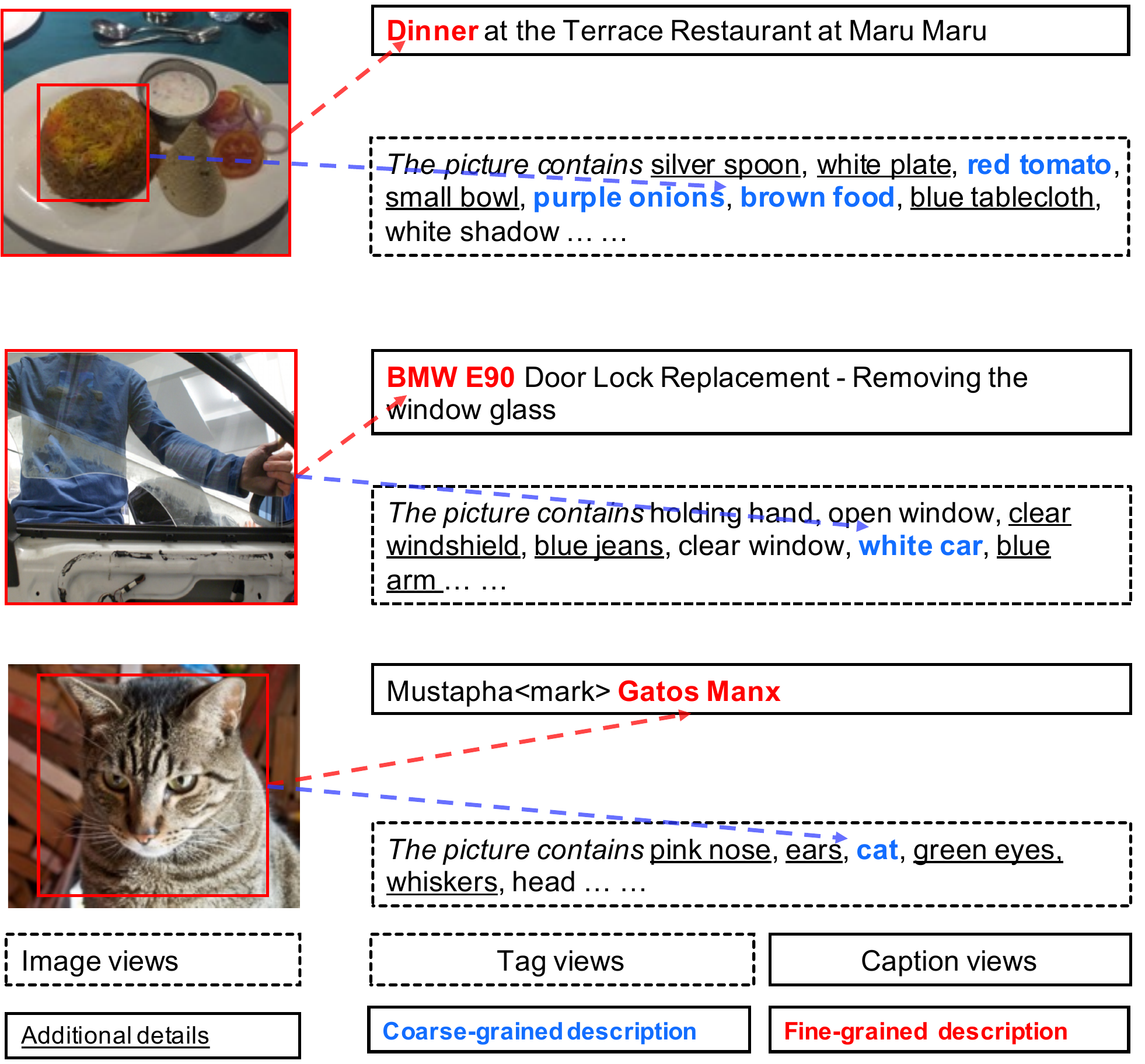}
\caption{The relations among different views: 1) Colored tokens share similar semantics, but they have different semantic granularity in captions (coarse-grained) and tags (fine-grained) view. 2) Underlined tokens in tag views provide missed information in the captions.}
\label{fig:case}
\end{figure}

To enhance the alignment across modalities on noisy image-text pairs, we introduce the special textual sequence (object tags) and apply the same text augmentation to get different views. Particularly, we construct the special textual sequence by concatenating a prompt (e.g., \textit{The picture contains}) and the attribute-object phrases extracted on raw images using a pre-trained object detector(e.g., BUTD \cite{Anderson2017up-down}). Besides, we keep these phrases for all the detected objects-attributes, corresponding with all the presented objects. We consider the special textual sequence holding more coarse-grained semantic textual units, which could be treated as a bridge between fine-grained (concrete) semantic units (e.g., named entities, low-frequency alternative name) in captions and abstract visual concepts in images, resulting in easing learning of cross-modal alignment. Besides, the model acquires more powerful image representative abilities due to the special textual sequence could provide information missed in captions. For a better interpretation of the relations among different views, we show some cases in Figure \ref{fig:case}.  

\subsection{Multi-view Contrastive Loss}
\label{subsec:Objectives}
For learning correlations of the views built in section \ref{subsec:Construction}, we propose a multi-view contrastive loss. We contrast two intra-modal view pairs and two inter-modal view pairs selected from constructed views at each iteration in a set $S$:  
$$S= \{(I_{v_1},I_{v_2}),(T_{v_1},T_{v_2}), (I_{v_1},T_{v_1}),(T_{v_1},I_{v_1})\}$$
For intra-modal pairs, $(I_{v_1},I_{v_2})$ and $(T_{v_1},T_{v_2})$ denote two visual views and textual views respectively. We select view pairs augmented on the same image or text as positives and others in the batch as negatives. For inter-modal pairs, $(I_{v_1},T_{v_1})$ and $(T_{v_1},I_{v_1})$ denote bi-directional visual/textual views pairs. Specifically, $T$ represents different textual views (caption or tag), and we randomly sample one of them (caption or tag) at each iteration. Then, the model will learn vision-language representation through contrasting the similarity of view pairs in $S$. Following InfoNCE \cite{Oord2018RepresentationLW}, we define the loss of each type of pair in $S$ as
\begin{equation}
L_{(x,y)}=-\frac{1}{N}\sum_{i}^{N}\log\frac{\exp( {h_{x}^{i}}^\top h_{y}^{i}/\tau)}{\sum_{j=1}^{N}\exp({h_{x}^{i}}^\top h_{y}^{j}/\tau)}
\end{equation}
where $x$ and $y$ denote different views in each pair of $S$, and $h$ denotes embeddings encoded by corresponding encoders, and $\tau$ is the temperature to scale the logits, and $N$ denotes the number of visual/textual views in the batch. Then, we minimize the overall contrastive loss for all view pairs with 

\begin{equation}
L_{multi-view}= \sum_{s \in S} \lambda_s L_{s}
\end{equation}
, where $\lambda$ are parameters to scale different losses, $s$ belongs to $S$ and is one type of view pairs in $S$.

\section{Experiment}
\label{sec:Experiment}
In this section, we present the pre-training datasets of our framework in Section \ref{sec:Pretraining Datasets} and settings in Section \ref{settings}, we evaluate our proposed model on cross-modal retrieval tasks in Section \ref{Cross_modal_Retrieval} and other vision-language tasks in Appendix \ref{Multi-modal_retrieval} and \ref{Transferring_fine-grained_understanding_tasks}, and conduct ablation studies in Section \ref{sec:alabtion}.
\subsection{Pre-training Datasets} 
\label{sec:Pretraining Datasets}
\paragraph{English pre-training datasets}  While many existing previous works incorporate massive private datasets such as 1.8B image-text pairs in ALIGN \cite{Jia2021ScalingUV}, we focus on utilizing only publicly available and fully out-of-domain datasets to build a comparable English benchmark for cross-modal retrieval. Thus, we built our pre-training data on three large-scale publicly available English datasets with 29M unique images: Conceptual Captions (CC) \cite{ng2020understanding}, Conceptual 12M \cite{changpinyo2021cc12m} and filtered Yahoo Flickr Creative Commons 100M \cite{Thomee2016YFCC100MTN} from \cite{Radford2021LearningTV} (remaining 14M). 
\paragraph{Chinese pre-training datasets} To generalize our framework to Chinese cross-modal retrieval, we scale the pre-training datasets to 1.5 billion image-text pairs consisting of the 1.1 billion web-crawled Chinese image-text pairs and 0.4 billion publicly available cross-modal datasets (CC12M, CC3M, YFCC, and LAION\cite{schuhmann2021laion400m} ). We collect 400M image-text pairs from the public datasets in the end, for some URLs were lost. We then translate translate the captions into Chinese by public machine translation system \footnote{We use Baidu Translation API Service at \url{https://fanyi.baidu.com/}}.
We present the details of the English and Chinese pre-training datasets in Table \ref{appendx_pretrain_data}.

\subsection{Pre-training Settings}
\label{settings}
For our English models, following previous method ALIGN \cite{Jia2021ScalingUV}, we consider EffcientNet-L2\cite{pmlr-v97-tan19a} as the image encoder and Bert-large \cite{devlin-etal-2019-bert} as the text encoder. For visual inputs, we resize the images to 256$\times$256 and implement random crop, horizontal flipping, jittering, gaussian blurring, and grayscale for image augmentation. We train the English model with a total batch size of 7168 on 112 A100 GPUs for 90K steps. We adopt Adam optimizer \cite{DBLP:journals/corr/KingmaB14} with a base learning rate of $10^{-4}$, which is warmed up linearly in 2K steps and decayed to $10^{-6}$ with a cosine schedule. 

We build our Chinese model following the similar settings except initializing the textual encoder with ERNIE \cite{sun2019ernie,sun2020ernie,sun2021ernie} and training with a total batch size of 23200. Besides, we explore another visual backbone widely used in VLP (i.e., Vision Transformer (ViT) \cite{dosovitskiy2021an}) as our image encoder (the details in Appendix \ref{overall_results} ).
\subsection{Results on Cross-modal Retrieval}
\label{Cross_modal_Retrieval}

\begin{table*}[]
\centering
\setlength{\tabcolsep}{0.6mm}{
\begin{tabular}{@{}llccccccccccccc@{}}
\toprule
\textbf{}                                &                                  &                               & \multicolumn{6}{c}{Flickr30K (1K test set)}                                                                                               & \multicolumn{6}{c}{MSCOCO (5K test set)}                                                                                 \\ \midrule
                                         & \multicolumn{1}{l|}{Method}      & \multicolumn{1}{c|}{\# Pre-train} & \multicolumn{3}{c|}{image $\to$ text}                                & \multicolumn{3}{c|}{text $\to$ image}                              & \multicolumn{3}{c|}{image $\to$ text}                              & \multicolumn{3}{c}{text $\to$ image}                \\
                                         & \multicolumn{1}{l|}{}            & \multicolumn{1}{c|}{images}   & R@1           & R@5            & \multicolumn{1}{c|}{R@10}           & R@1           & R@5           & \multicolumn{1}{c|}{R@10}          & R@1           & R@5           & \multicolumn{1}{c|}{R@10}          & R@1                 & R@5           & R@10          \\ \midrule
 & \multicolumn{1}{l|}{Unicoder-VL} & \multicolumn{1}{c|}{4M}       & 48.4          & 76.0           & \multicolumn{1}{c|}{85.2}           & 64.3          & 85.8          & \multicolumn{1}{c|}{92.3}          & 44.9          & 71.2          & \multicolumn{1}{c|}{80.4}          & 32.3                & 59.0          & 70.2          \\
\multicolumn{1}{c}{}                    & \multicolumn{1}{l|}{UNITER}      & \multicolumn{1}{c|}{4M}       & 83.6          & 95.7           & \multicolumn{1}{c|}{97.7}           & 68.7          & 89.2          & \multicolumn{1}{c|}{93.9}          & -             & -             & -                                  & -                   & -             & -             \\
\multicolumn{1}{c}{}                    & \multicolumn{1}{l|}{CLIP}        & \multicolumn{1}{c|}{400M}     & 88.0          & 98.7           & \multicolumn{1}{c|}{99.4}           & 68.7          & 90.6          & \multicolumn{1}{c|}{95.2}          & 58.4          & 81.5          & \multicolumn{1}{c|}{88.1}          & 37.8                & 62.4          & 72.2          \\
\multicolumn{1}{c}{}                    & \multicolumn{1}{l|}{ALIGN}       & \multicolumn{1}{c|}{1.8B}     & 88.6          & 98.7           & \multicolumn{1}{c|}{99.7}           & 75.7          & 93.8          & \multicolumn{1}{c|}{96.8}          & 58.6          & 83.0          & \multicolumn{1}{c|}{89.7}          & 45.6                & 69.8          & 78.6          \\
\multicolumn{1}{c}{}                    & \multicolumn{1}{l|}{FILIP}       & \multicolumn{1}{c|}{340M}     & 89.8          & 99.2           & \multicolumn{1}{c|}{99.8}           & 75.0          & 93.4          & \multicolumn{1}{c|}{96.3}          & 61.3          & 84.3          & \multicolumn{1}{c|}{90.4}          & 45.9       & 70.6          & 79.3          \\
\multicolumn{1}{c}{}                    &\multicolumn{1}{l|}{Florence} & \multicolumn{1}{c|}{900M} & 90.9 & 99.1 & \multicolumn{1}{c|}{-} & 76.7 & 93.6 & \multicolumn{1}{c|}{-} & 64.7 & 85.9 & \multicolumn{1}{c|}{-} & 47.2 & 71.4 & - \\
\multicolumn{1}{c}{}                    &\multicolumn{1}{l|}{FLAVA} & \multicolumn{1}{c|}{70M} & 67.7 & 94.0 & \multicolumn{1}{c|}{-} & 65.2 & 89.4 & \multicolumn{1}{c|}{-} & 42.7 & 76.8 & \multicolumn{1}{c|}{-} & 38.4 & 67.5 & - \\
\multicolumn{1}{c}{}                    &\multicolumn{1}{l|}{CoCa$\dagger$} & \multicolumn{1}{c|}{4.8B} & \textbf{92.5} & \textbf{99.5} & \multicolumn{1}{c|}{\textbf{99.9}} & \textbf{80.4} & \textbf{95.7} & \multicolumn{1}{c|}{\textbf{97.7}} & \textbf{66.3} & \textbf{86.2 }& \multicolumn{1}{c|}{\textbf{91.8}} & \textbf{51.2} & \textbf{74.2} &\textbf{ 82.0} \\

\multicolumn{1}{c}{}                    & \multicolumn{1}{l|}{ERNIE-ViL 2.0}      & \multicolumn{1}{c|}{29M}      & {91.2} & {99.1} & \multicolumn{1}{c|}{{99.8}} & { 77.4} & {93.8} & \multicolumn{1}{c|}{{96.4}} & {63.1} & {85.7} & \multicolumn{1}{c|}{{91.4}} & {46.0} & {71.4} & {80.4} \\ \cmidrule(l){2-15} 
\multicolumn{1}{c}{}                    & \multicolumn{1}{l|}{FILIP*}      & \multicolumn{1}{c|}{340M}     & 95.4          & 99.8           & \multicolumn{1}{c|}{100.0}          & 84.7          & 97.0          & \multicolumn{1}{c|}{98.7}          & -             & -             & \multicolumn{1}{c|}{-}             & -                   & -             & -             \\
\multicolumn{1}{c}{}                    & \multicolumn{1}{l|}{ALBEF*}      & \multicolumn{1}{c|}{4M}       & 94.1          & 99.5           & \multicolumn{1}{c|}{99.7}           & 82.8          & 96.3          & \multicolumn{1}{c|}{98.1}          & -             & -             & \multicolumn{1}{c|}{-}             & -                   & -             & -             \\
\multicolumn{1}{c}{}                    & \multicolumn{1}{l|}{ERNIE-ViL 2.0*}     & \multicolumn{1}{c|}{29M}      & \textbf{96.1} & \textbf{99.9}  & \multicolumn{1}{c|}{\textbf{100.0}} & \textbf{85.0} & \textbf{97.0} & \multicolumn{1}{c|}{\textbf{98.3}} & -             & -             & \multicolumn{1}{c|}{-}             & -                   & -             & -            
     \\ \bottomrule
\end{tabular}
}
\caption{ Zero-shot English cross-modal retrieval results on Flickr30K and MSCOCO datasets, compared with previous best VLP models with or without in-domain datasets based on different architecture. *: the model fine-tuned on MSCOCO dataset. CoCa$\dagger$ uses 165x more image-text pairs.}
\label{table:crossretrieval_zero-shot}
\end{table*}

\begin{table*}[]
\centering
\setlength{\tabcolsep}{0.6mm}{
\begin{tabular}{@{}llccccccccccccc@{}}
\toprule
\textbf{}                                &                                  &                               & \multicolumn{6}{c}{Flickr30K (1K test set)}                                                                                               & \multicolumn{6}{c}{MSCOCO (5K test set)}                                                                                 \\ \midrule
                                         & \multicolumn{1}{l|}{Method}      & \multicolumn{1}{c|}{\# Pre-train} & \multicolumn{3}{c|}{image $\to$ text}                                & \multicolumn{3}{c|}{text $\to$ image}                              & \multicolumn{3}{c|}{image $\to$ text}                              & \multicolumn{3}{c}{text $\to$ image}                \\
                                         & \multicolumn{1}{l|}{}            & \multicolumn{1}{c|}{images}   & R@1           & R@5            & \multicolumn{1}{c|}{R@10}           & R@1           & R@5           & \multicolumn{1}{c|}{R@10}          & R@1           & R@5           & \multicolumn{1}{c|}{R@10}          & R@1                 & R@5           & R@10          \\ \midrule

 & \multicolumn{1}{l|}{UNITER}      & \multicolumn{1}{c|}{4M}       & 87.3          & 98.0           & \multicolumn{1}{c|}{99.2}           & 75.6          & 94.1          & \multicolumn{1}{c|}{96.8}          & 65.7          & 88.6          & \multicolumn{1}{c|}{93.8}          & 52.9                & 79.9          & 88.0          \\
\multicolumn{1}{l}{}                    & \multicolumn{1}{l|}{OSCAR}       & \multicolumn{1}{c|}{4M}       & -             & -              & \multicolumn{1}{c|}{-}              & -             & -             & \multicolumn{1}{c|}{-}             & 73.5          & 92.2          & \multicolumn{1}{c|}{96.0}          & 57.5                & 82.8          & 89.8          \\
\multicolumn{1}{l}{}                    & \multicolumn{1}{l|}{ERNIE-ViL}   & \multicolumn{1}{c|}{4M}       & 88.7          & 97.3           & \multicolumn{1}{c|}{99.1}           & 75.1          & 93.4          & \multicolumn{1}{c|}{96.3}          & -             & -             & \multicolumn{1}{c|}{-}             & -                   & -             & -             \\
\multicolumn{1}{l}{}                    & \multicolumn{1}{l|}{ALIGN}       & \multicolumn{1}{c|}{1.8B}     & 95.3          & 99.8           & \multicolumn{1}{c|}{100.0}          & 84.9          & 97.4          & \multicolumn{1}{c|}{98.6}          & 77.0          & 93.5          & \multicolumn{1}{c|}{96.9}          & 59.9                & 83.3          & 89.8          \\
\multicolumn{1}{l}{}                    & \multicolumn{1}{l|}{ALBEF}       & \multicolumn{1}{c|}{14M}      & 95.9          & 99.8           & \multicolumn{1}{c|}{100.0}          & 85.6          & 97.5          & \multicolumn{1}{c|}{98.9}          & 77.6          & 94.3          & \multicolumn{1}{c|}{97.2}          & 60.7                & 84.3          & 90.5          \\
\multicolumn{1}{l}{}                    & \multicolumn{1}{l|}{FILIP}       & \multicolumn{1}{c|}{340M}     & 96.6          & 100.0          & \multicolumn{1}{c|}{100.0}          & 87.1          & 97.7          & \multicolumn{1}{c|}{99.1}          & 78.9        & 94.4        & \multicolumn{1}{c|}{97.4}          & 61.2                & 84.3          & 90.5         \\
\multicolumn{1}{l}{}                    & \multicolumn{1}{l|}{Florence}       & \multicolumn{1}{c|}{900M}     & \textbf{97.2}         & 99.9          & \multicolumn{1}{c|}{-}          & 87.9          & 98.1       & \multicolumn{1}{c|}{-}          & \textbf{81.8}         & \textbf{95.2}     & \multicolumn{1}{c|}{\textbf{-}}          & \textbf{63.2}             & \textbf{85.7}  & \textbf{-  }        \\
\multicolumn{1}{l}{}                    & \multicolumn{1}{l|}{ERNIE-ViL 2.0}      & \multicolumn{1}{c|}{29M}      & \textbf{97.2} & \textbf{100.0} & \multicolumn{1}{c|}{\textbf{100.0}} & \textbf{93.3} & \textbf{99.4} & \multicolumn{1}{c|}{\textbf{99.8}} & 77.4          & 93.6          & \multicolumn{1}{c|}{97.1}          & 59.5                & 83.4          & 90.1        \\ \bottomrule
\end{tabular}
}
\caption{ Fine-tuned English cross-modal retrieval results on Flickr30K and MSCOCO datasets, compared with previous best VLP models with or without in-domain datasets based on different architecture.}
\label{table:crossretrieval}
\end{table*}

\begin{table*}[]
\centering
\setlength{\tabcolsep}{0.6mm}{
\begin{tabular}{@{}lcccccccccccccc@{}}
\toprule
 & \multicolumn{7}{c}{AIC-ICC} & \multicolumn{7}{c}{COCO-CN} \\ \midrule
\multicolumn{1}{l|}{Models} & \multicolumn{3}{c|}{image $\to$ text} & \multicolumn{3}{c|}{text$\to$image} & \multicolumn{1}{c|}{mean} & \multicolumn{3}{c|}{image$\to$text} & \multicolumn{3}{c|}{text$\to$image} & mean \\
\multicolumn{1}{l|}{} & R@1 & R@5 & \multicolumn{1}{c|}{R@10} & R@1 & R@5 & \multicolumn{1}{c|}{R@10} & \multicolumn{1}{c|}{Recall} & R@1 & R@5 & \multicolumn{1}{c|}{R@10} & R@1 & R@5 & \multicolumn{1}{c|}{R@10} & Recall \\ \midrule
\multicolumn{1}{l|}{BriVL} & 20.3 & 37.0 & \multicolumn{1}{c|}{45.6} & 14.4 & 30.4 & \multicolumn{1}{c|}{39.1} & \multicolumn{1}{c|}{31.1} & - & - & \multicolumn{1}{c|}{-} & - & - & \multicolumn{1}{c|}{-} & - \\
\multicolumn{1}{l|}{Taiyi} & -& - & \multicolumn{1}{c|}{-} & - & - & \multicolumn{1}{c|}{-} & \multicolumn{1}{c|}{-} & - & - & \multicolumn{1}{c|}{-} & 51.5 & 81.0 & \multicolumn{1}{c|}{90.4} & - \\
\multicolumn{1}{l|}{R2D2} & 30.7 & 47.2 & \multicolumn{1}{c|}{52.9} & 14.9 & 28.1 & \multicolumn{1}{c|}{33.4} & \multicolumn{1}{c|}{34.5} & 63.3 & 89.3 & \multicolumn{1}{c|}{95.7} & 56.4 & 85.0 & \multicolumn{1}{c|}{93.1} & 80.5 \\
\multicolumn{1}{l|}{Wukong} & - & - & \multicolumn{1}{c|}{-} & - & - & \multicolumn{1}{c|}{-} & \multicolumn{1}{c|}{-} & 55.2 & 81.0 & \multicolumn{1}{c|}{90.6} & 53.4 & 80.2 & \multicolumn{1}{c|}{90.1} & 75.1 \\
\multicolumn{1}{l|}{M3P} & - & - & \multicolumn{1}{c|}{-} & - & - & \multicolumn{1}{c|}{-} & \multicolumn{1}{c|}{-} & - & - & \multicolumn{1}{c|}{-} & - & - & \multicolumn{1}{c|}{-} & 32.3 \\
\multicolumn{1}{l|}{ERNIE-ViL 2.0} & \textbf{33.7} & \textbf{52.1} & \multicolumn{1}{c|}{\textbf{60.0}} & \textbf{19.0} & \textbf{35.3} & \multicolumn{1}{c|}{\textbf{43.5}} & \multicolumn{1}{c|}{\textbf{40.6}} & \textbf{69.1} & \textbf{92.9} & \multicolumn{1}{c|}{\textbf{97.1}} & \textbf{69.6} & \textbf{91.2} & \multicolumn{1}{c|}{\textbf{96.9}} & \textbf{86.1} \\ \bottomrule
\end{tabular}
}
\caption{ Zero-shot Chinese cross-modal retrieval results on COCO-CN and AIC-ICC datasets, compared with the previous Chinese VLP models.}
\label{table_ch_1}
\end{table*}
Under the zero-shot setting, we directly use the pre-trained model for image-text retrieval. In the fine-tuning phase, for a fair comparison with previous VL pre-training methods, we only use the conventional views (image-caption pairs). To evaluate text-image retrieval, we adopt recall at K (R@K) as the metric. 
\paragraph{English cross-modal retrieval}
We evaluate our English pre-trained model for cross-modal retrieval on Flickr30K \cite{flickrentitiesijcv} and MSCOCO \cite{Chen2015MicrosoftCC} in zero-shot and fine-tuning settings. Following the widely used splits in \cite{Karpathy2017DeepVA}, we split MSCOCO and Flickr30K into the train/validation/test set with 113K/5K/5K and 29K/1K/1K, respectively. As shown in Table \ref{table:crossretrieval_zero-shot} and Table \ref{table:crossretrieval}, ERNIE-ViL 2.0 achieves a comparable results compared with previous works. Specifically, compared to methods \cite{chen2020uniter,li2020unicoder,singh2021flava,Yu2021ERNIEViLKE,Li2021AlignBF,Li2020OscarOA} only using publicly available datasets, ERNIE-ViL 2.0 obtain significant improvements both in zero-shot and fine-tuning settings. For the works \cite{Radford2021LearningTV,Jia2021ScalingUV,DBLP:journals/corr/abs-2111-07783,yuan2021florence,yu2022coca} using large-scale private datasets, we also present superior results of zero-shot cross-modal retrieval except for CoCa \cite{yu2022coca} with 165x more image-text pairs. Notably, we outperform ALIGN \cite{Jia2021ScalingUV} with the same model size and architecture using 62x fewer pre-training image-text pairs.

\paragraph{Chinese cross-modal retrieval}
We evaluate our Chinese pre-training models on two Chinese cross-modal benchmarks COCO-CN \cite{li2018cococn} and AIC-ICC \cite{wu2019large}, where we use standard 1K test splits for COCO-CN and the validation split for AIC-ICC (the first 10K images, following BriVL \cite{Huo2021WenLanBV}). As shown in Table \ref{table_ch_1}, ERNIE-ViL 2.0 outperforms all existing methods \cite{Huo2021WenLanBV,Fengshenbang-LM,xie2022zero,gu2022wukong,M3P,Radford2018ImprovingLU} and achieves SOTA results both on COCO-CN and AIC-ICC with zero-shot settings.  Compared to previous best methods \cite{xie2022zero}, ERNIE-ViL 2.0 obtain an improvements of 6.1\% meanRecall on AIC-ICC and 5.6\% meanRecall on COCO-CN. Besides, we also present the results of the models with ViT as visual encoder in Appendix Table \ref{details_results}.

\subsection{Ablation Study}
\begin{figure*}[]
\centering
\includegraphics[width=5in]{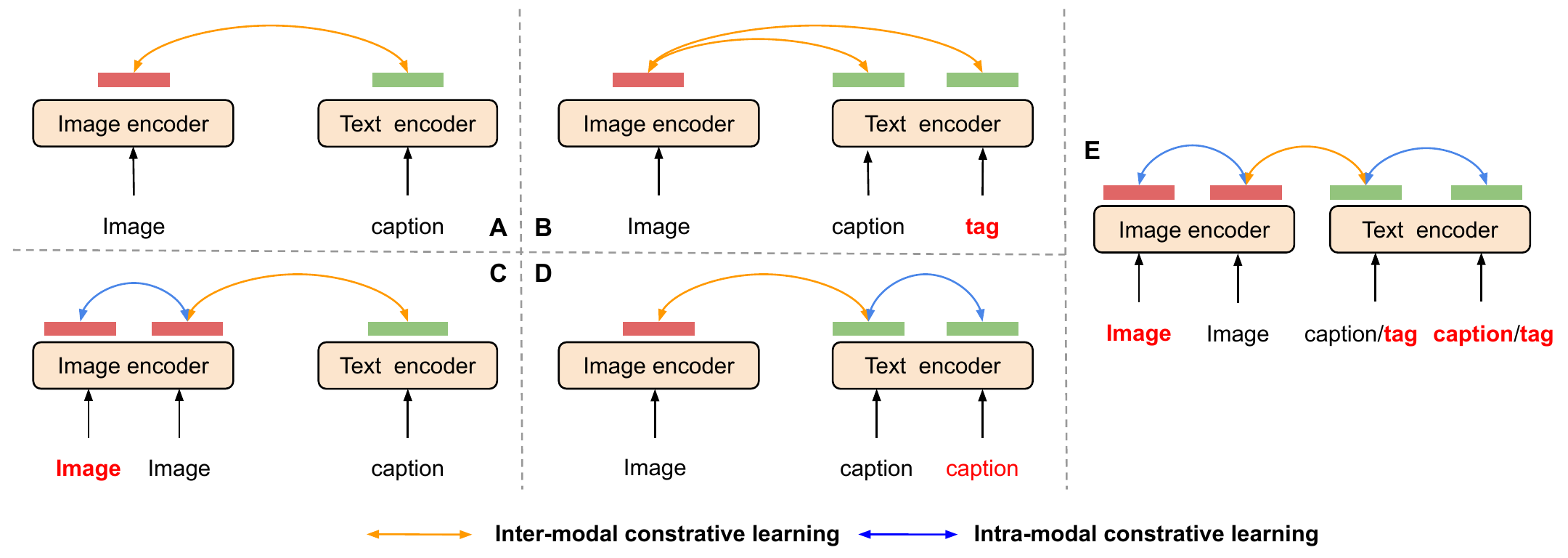}
\caption{Ablation study with five experiments. A (baseline): conventional single-view contrastive learning. B integrates image-tags (inter-modal) view pairs into baseline A. C and D integrate image-image view pairs and caption-caption view pairs (intra-modal) into baseline A, respectively. E merges entire strategies (A+B+C+D), namely ERNIE-ViL 2.0. }
\label{ablation_1}
\end{figure*} 
\begin{table*}[]
\centering
\begin{tabular}{lll|c|c|c}
\toprule
No & Model & Types of view pairs & Image Retrieval & Text Retrieval &  \\ 
 &  &  & mR & mR & mR\\ \midrule
A & Baseline & \multirow{2}{*}{Inter-modal} & 81.9 & 91.6& 86.8  \\
B & + (tag,image) &  & 83.7 & 92.6& 88.2 \\ \midrule
C & + (image,image) & \multirow{2}{*}{Intra-modal} & 83.6 & 91.3 & 87.5\\
D & + (caption,caption) &  & 83.8 & 92.1& 88.0\\ \midrule
E & ERNIE-ViL2.0$_{ablation}$ & Inter-modal \& Inter-modal & 85.1 & 93.0 &89.1\\
\bottomrule
\end{tabular}
\caption{ The results of the ablated models to study the effect of our proposed strategies. $(X,Y)$ donates contrasting view pairs (e.g., (tag,image) is learn inter-modal correlations between tags and images). The metric of all experiments is mR (meanRecall), which is the average of R@1, R@5 and R@10.
ERNIE-ViL 2.0$_{ablation}$ uses full purposed strategies of ERNIE-ViL 2.0 with the ablation setting. }
\label{ablation_2}
\end{table*}
\label{sec:alabtion}
We study the benefits of the additional visual and textual views enhancing the cross-modal contrastive learning on the English pre-training datasets. Figure \ref{ablation_1} illustrates the detailed contrastive models used in our ablation experiments. We treat the conventional single-view constrative learning method as our baseline (model A). We incorporate intra-modal views for contrastive learning with model C for visual views and model D for textual views. We also try adding another inter-modal views for contrasting using the sequence of tags as the special textual view (model B). Finally, we arrive at our final multi-view contrastive learning model E. 
We pre-train all the ablated models using CC \cite{ng2020understanding} and CC12M \cite{changpinyo2021cc12m} as the pre-training datasets for 50,000 steps with a batch size of 1024. All the models follow the same architecture and other hyper-parameters with our final ERNIE-ViL 2.0 model. We measure the performance of the models on zero-shot cross-modal retrieval of Flickr30K and use meanRecall as the evaluation metric.

The results of the experiments are listed in Table \ref{ablation_2}. We  obtain the following observations from the performance comparisons.

\noindent\textbf{Adding intra-modal views} While the conventional contrastive learning method use the image-text pair, rely only on the cross-modal view pairs for contrastive learning. Incorporating the intra-modal views pairs, both visual view pairs and textual view pairs,  brings improvement for image-text retrieval task. Specifically, adding text-text pairs for contrasting obtains an absolute improvement 1.2\% on meanRecall  (Model D compared to Model A) while adding image-image pairs brings an absolute improvement of 0.7\% (Model C compared to Model A). 

\noindent\textbf{Enhancing inter-modal views} 
We consider the sequence of tags as special textual view and construct a new inter-modal view pairs. We observe that with this type of enhanced inter-modal view pairs, the contrastive learning learning better representations for image and text. Adding (tag, image) pairs for contrasting brings a significant absolute improvement of 1.6\% on meanRecall (Model B compared to Model A). 
Actually, the sequence of tags serves as a bridge between fine-grained semantic units in captions and abstract visual concepts in images, resulting in easing learning of cross-modal alignment. 

\noindent\textbf{Multi-view contrastive learning} Bringing the enhanced inter-modal views and intra-modal views together, we arrive at the final multi-view contrastive learning method, ERNIE-ViL 2.0. Compared to the image-text contrastive methods, ERNIE-ViL 2.0 makes a signifcant absolute improvement of 2.3\% on meanRecall for image-text retrieval. For text-to-image retrieval the absolute improvement (3.2\%) on meanRecall is more significant given that the text-to-image task is more challenging (there is five ground-truth captions for each image and only one ground-truth for each image), demonstrating the effectiveness of our multi-view contrastive learning framework.

\section{Analysis of Learned Embeddings}
\begin{figure}[]
\centering
\includegraphics[width=6in]{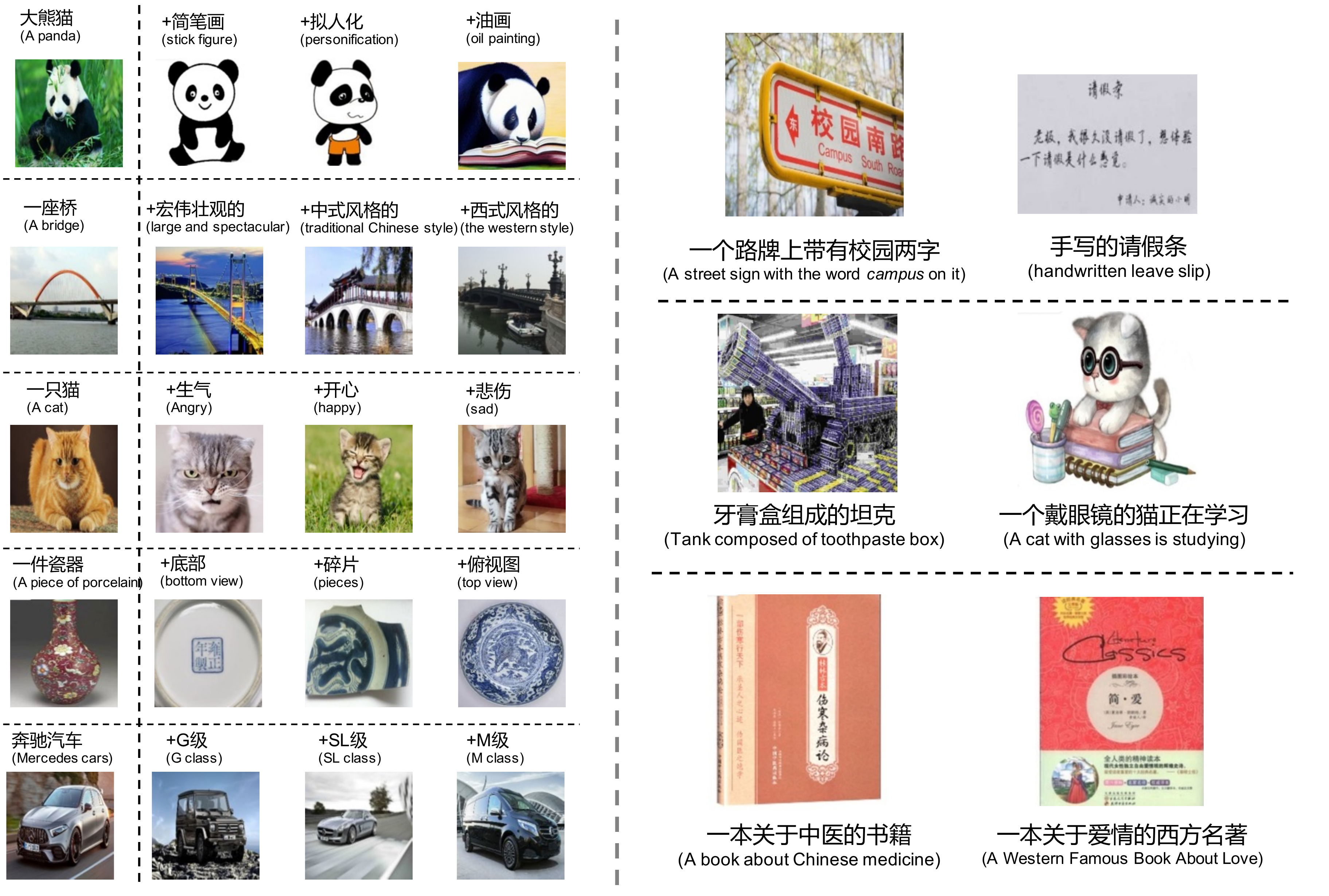}
\caption{Image retrieval with fine-grained text queries using ERNIE-ViL 2.0's embeddings. The left side lists general descriptions with qualifying words (i.e., style, viewpoint, class, etc.). The right side lists the complex queries (i.e., The first row requires recognizing the detailed character in images for given questions. The second row presents some long-tail queries. The third row shows queries containing abstract concepts.) }
\label{case_t2i_1}
\end{figure} 
We construct a Chinese image-text retrieval system and perform image retrieval and text retrieval to study the learned embeddings by ERNIE-ViL 2.0 (Chinese model).

\paragraph{Image retrieval} We perform image retrieval on 300K web-crawled images separated from our training data. Figure \ref{case_t2i_1} shows the top-1 text-to-image retrieval results with two categories of manual queries. The first category of queries consists of general words with different qualifying words(e.g., personification style, animal facial expression, the levels for the specified cars), and ERNIE-ViL 2.0 can retrieve the precise image given the detailed description. Besides, the second part (right side in Figure \ref{case_t2i_1}) presents retrieval results with several hard queries (e.g., fine-grained words, abstract concepts, long-tail phrases, etc.) validating the superior capability of our models.

\begin{figure*}[]
\centering
\includegraphics[width=6in]{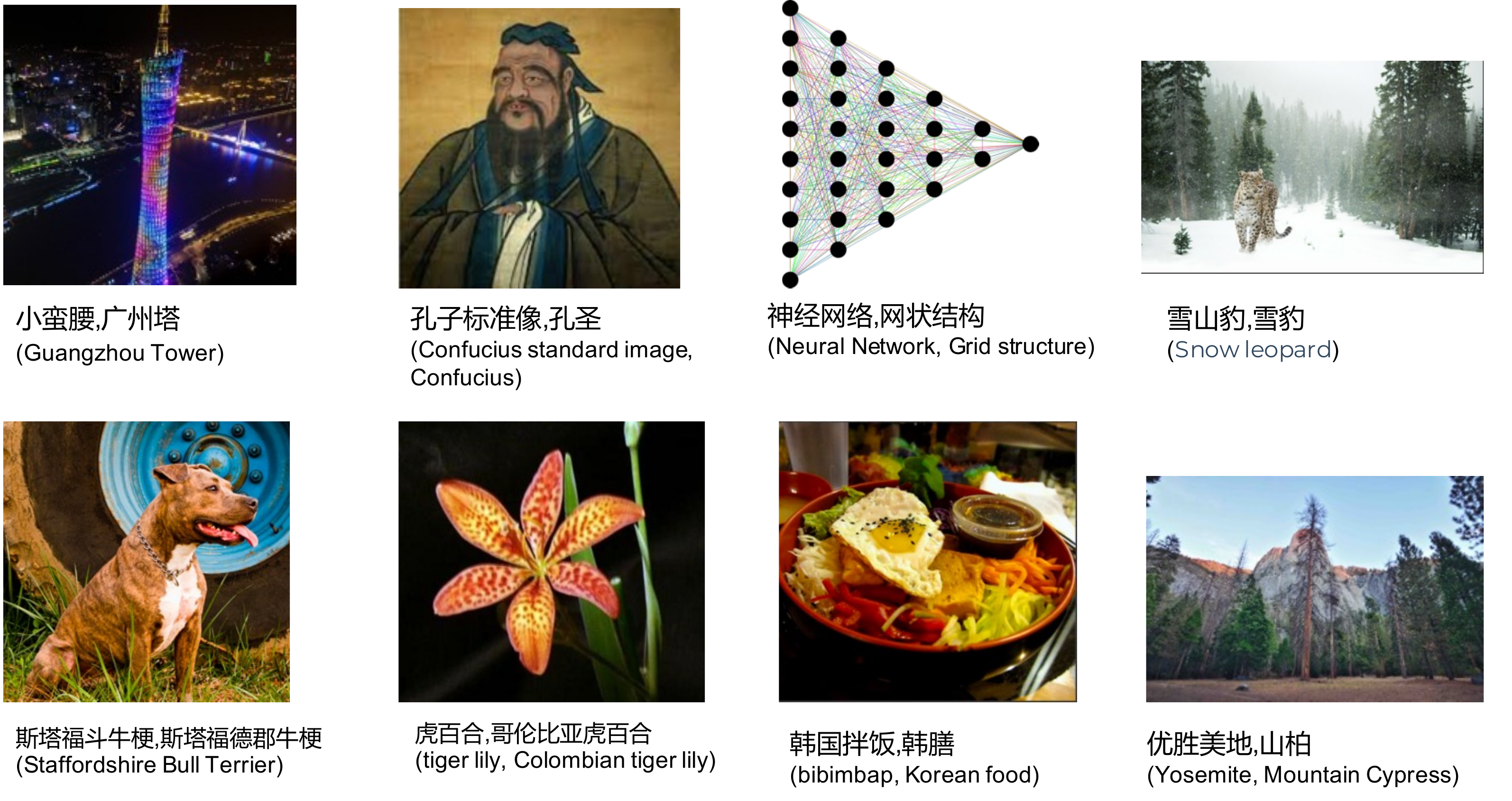}
\caption{The text retrieval with images from different domains (e.g., Landmark, famous people, artworks, flowers, pets, foods, etc.) using ERNIE-ViL 2.0's embeddings.}
\label{case_I2T_1}
\end{figure*}

\paragraph{Text retrieval}  We perform text retrieval on 41k tags collected from a Chinese public lexicon (e.g.,THUOCL\footnote{http://thuocl.thunlp.org/}). We present top-2 retrieval results for each image selected from different domains (e.g., food, landmark, pets, and famous people). As shown in Figure \ref{case_I2T_1}, ERNIE-ViL 2.0 has a powerful ability to recognize the fine-grained concepts and instances in images. As mentioned in Section \ref{subsec:Objectives}, we consider ERNIE-ViL 2.0 can easily learn such fine-grained concepts on noisy cross-modal data leveraging construct the contrastive learning between object tags and images.

\section{Conclusion}
We propose a Multi-view contrastive learning framework ERNIE-ViL 2.0 for cross-modal retrieval, to improve the robustness and generalization of the cross-modal representation through incorporating diverse views into the contrastive learning framework. ERNIE-ViL 2.0 attempts to build both inter-modal and intra-modal correlations simultaneously. Moreover, we construct a special textual view as a bridge to ease the learning of the challenging semantic alignment between noisy image-text pairs. ERNIE-ViL 2.0 achieves SOTA results on Chinese cross-modal retrieval tasks, and comparable results on English cross-modal retrieval pre-trained with fewer image-text pairs and hence build a solid and robust benchmark with only publicly available datasets. For future work, we will study multi-view learning for the cross-encoder architecture, which obtain better results for complex Vision-Language reasoning tasks. 

\bibliographystyle{unsrt}  
\bibliography{references}  
\newpage

\appendix
\section{The Details of Pre-training Datasets}
We list the details of pre-training datasets used in ERNIE-ViL 2.0 in Table \ref{appendx_pretrain_data}, where ERNIE-ViL$_{1.1B}$ denotes 1.1 billion image-text pairs crawled from the Chinese websites.
\begin{table*}[]
\centering
\begin{tabular}{ccc}
\toprule
Language & Source & \# Image \\ \midrule
English & CC3M, CC12M, YFCC & 29M \\
Chinese & CC3M, CC12M, YFCC,LAION, ERNIE-ViL$_{1.1B}$ & 1.5B \\ \bottomrule
\end{tabular}
\caption{The details of pre-training datasets used in ERNIE-ViL 2.0}
\label{appendx_pretrain_data}
\end{table*}
\section{Multi-modal Retrieval}
\label{Multi-modal_retrieval}
Due to that our framework learns intra-modal and inter-modal views simultaneously, we consider another datasets Crisscrossed Caption (CxC) \cite{parekh2020crisscrossed}, which contains inter-modal and intra-modal retrieval tasks. CxC extends MSCOCO datasets to 267,095 caption-caption, image-image, and caption-image pairs. Specifically, following \cite{parekh2020crisscrossed}, we construct positives and negtives in evaluation for  semantic textual similarity (STS), semantic image-text similarity (SITS) and semantic image similarity (SIS). We evaluate ERNIE-ViL 2.0 (English model) on CXC using the fine-tuned model on COCO datasets. The Table \ref{Multimodalretrieval} reports experimental results, showing that ERNIE-ViL 2.0 outperforms ALIGN (previous best method) both in image-to-text and text-to-text retrieval. 

\begin{table*}[]
\centering
\begin{tabular}{lllll}
\toprule
    & I2T  & T2I  & T2T  & I2I  \\ \midrule
DET2T+I2T     & 77.3 & 65.7 & 60.4 & 65.7 \\
ALIGN         & 89.9 & \textbf{79.2} & 62.5 & \textbf{73.3} \\
ERNIE-ViL 2.0 & \textbf{90.0} & 79.0 & \textbf{64.7} & 72.4 \\ \bottomrule
\end{tabular}
\caption{Multimodal retrieval results at Crisscrossed Captions (CxC), compared with multi-modal methods: ALIGN \cite{Jia2021ScalingUV}, DET2T+I2T \cite{parekh2020crisscrossed}. The metric is the meanRecall (mR). I2T: text $\to$ image, I2T:image $\to$ text, I2I:image $\to$ image, T2T: text $\to$ text }
\label{Multimodalretrieval}
\end{table*}

\section{Transferring to Fine-grained Understanding Tasks}
\label{Transferring_fine-grained_understanding_tasks}
For further evaluating the transferability of ERNIE-ViL 2.0, we transfer our English pre-trained model to fine-grained multimodal tasks. We consider two tasks: SNLI-VE \cite{xie2019visual} and NLVR$^2$ \cite{suhr-etal-2019-corpus}. SNLI-VE is a visual entailment task to predict whether a given image entails a text. NLVR$^2$ requires the model to predict the relations between a given sentence and an image pair. Following UNITER's \cite{chen2020uniter} settings, classification accuracy is adopted to measure two benchmarks (SNLI-VE and NLVR$^2$). Our model predicts the class probabilities on SNLI-VE using a multi-layer perception (MLP) on a concatenated representation of image and text encoder outputs. 
For NLVR$^2$, after encoding the two image into individual embedding, we concatenate and feed them to a MLP layer to obtain the visual embedding. The textual embedding and the visual embedding are concatenated and fed into the classifier to get the final scores. 
We present the results in Table \ref{fine-vl}. Although those tasks need more fine-grained interaction of visual/textual features, our method still presents comparable results with a global cross-modal representation.
 \begin{table*}
\centering
\begin{tabular}{@{}l|ll@{}}
\toprule
& SNLI-VE & NLVR$^2$ \\ \midrule
UNITER          & 78.3    & -     \\
VisualBERT      & -       & 67.0  \\
ERNIE-ViL 2.0 & 76.6    & 66.0  \\ \bottomrule
\end{tabular}
\caption{Compared with previous methods with Fine-grained vision-language interaction on V+L downstream tasks: SNLI-VE and NLVR2, ERNIE-ViL 2.0 present comparable results with shallow cross-modal interaction }
\label{fine-vl}
\end{table*}
\section{The Overall Results of Chinese Cross-modal Retrieval }
\label{overall_results}
We list the overall results of Chinese cross-modal results on AIC-ICC and COCO-CN with different architecture in Table \ref{details_results}
\begin{table}[]

\centering

\begin{tabular}{@{}cccccc@{}}
\toprule
\multicolumn{1}{l}{Models} & \multicolumn{2}{l}{} & \multicolumn{1}{l}{ERNIE-ViL 2.0$_{large}$} & \multicolumn{1}{l}{ERNIE-ViL 2.0$_{large}$} & \multicolumn{1}{l}{ERNIE-ViL 2.0$_{base}$} \\ \midrule
\multicolumn{1}{l|}{Backbone} & \multicolumn{2}{l|}{Image Encoder} & \multicolumn{1}{l|}{ViT-L-14} & \multicolumn{1}{l|}{EffcientNet-L2} & \multicolumn{1}{l}{ViT-B-16} \\ \cmidrule(l){2-6} 
\multicolumn{1}{l|}{} & \multicolumn{2}{l|}{Text    Encoder} & \multicolumn{1}{l|}{ERNIE-Large} & \multicolumn{1}{l|}{ERNIE-Large} & \multicolumn{1}{l}{ERNIE-Base}  \\ \midrule
\multicolumn{1}{l|}{} & \multicolumn{2}{l|}{Batch size} & \multicolumn{1}{c|}{8000} & \multicolumn{1}{c|}{23200} & \multicolumn{1}{c}{14000}  \\ \midrule
\multicolumn{1}{c|}{\multirow{7}{*}{AIC-ICC}} & \multirow{3}{*}{image$\to$text} & \multicolumn{1}{c|}{R@1} & \multicolumn{1}{c|}{32.3} & \multicolumn{1}{c|}{\textbf{33.7}} & \multicolumn{1}{c}{30.4}  \\
\multicolumn{1}{c|}{} &  & \multicolumn{1}{c|}{R@5} & \multicolumn{1}{c|}{51.6} & \multicolumn{1}{c|}{\textbf{52.1}} & \multicolumn{1}{c}{48.9}  \\
\multicolumn{1}{c|}{} &  & \multicolumn{1}{c|}{R@10} & \multicolumn{1}{c|}{59.9} & \multicolumn{1}{c|}{\textbf{60.0}} & \multicolumn{1}{c}{57.5}  \\ \cmidrule(l){2-6} 
\multicolumn{1}{c|}{} & \multirow{3}{*}{text$\to$image} & \multicolumn{1}{c|}{R@1} & \multicolumn{1}{c|}{\textbf{20.2}} & \multicolumn{1}{c|}{19.0} & \multicolumn{1}{c}{17.9}  \\
\multicolumn{1}{c|}{} &  & \multicolumn{1}{c|}{R@5} & \multicolumn{1}{c|}{\textbf{37.0}} & \multicolumn{1}{c|}{35.3} & \multicolumn{1}{c}{34.2}  \\
\multicolumn{1}{c|}{} &  & \multicolumn{1}{c|}{R@10} & \multicolumn{1}{c|}{\textbf{45.4}} & \multicolumn{1}{c|}{43.5} & \multicolumn{1}{c}{42.5}   \\ \midrule
\multicolumn{1}{c|}{\multirow{7}{*}{COCO-CN}} & \multirow{3}{*}{image$\to$text} & \multicolumn{1}{c|}{R@1} & \multicolumn{1}{c|}{68.6} & \multicolumn{1}{c|}{\textbf{69.1}} & \multicolumn{1}{c}{66.5} \\
\multicolumn{1}{c|}{} &  & \multicolumn{1}{c|}{R@5} & \multicolumn{1}{c|}{92.5} & \multicolumn{1}{c|}{\textbf{92.9}} & \multicolumn{1}{c}{91.6}  \\
\multicolumn{1}{c|}{} &  & \multicolumn{1}{c|}{R@10} & \multicolumn{1}{c|}{\textbf{97.5}} & \multicolumn{1}{c|}{97.1} & \multicolumn{1}{c}{96.2}  \\ \cmidrule(l){2-6} 
\multicolumn{1}{c|}{} & \multirow{3}{*}{text$\to$image} & \multicolumn{1}{c|}{R@1} & \multicolumn{1}{c|}{\textbf{70.1}} & \multicolumn{1}{c|}{69.6} & \multicolumn{1}{c}{65.9} \\
\multicolumn{1}{c|}{} &  & \multicolumn{1}{c|}{R@5} & \multicolumn{1}{c|}{\textbf{92.5}} & \multicolumn{1}{c|}{91.2} & \multicolumn{1}{c}{90.1}  \\
\multicolumn{1}{c|}{} &  & \multicolumn{1}{c|}{R@10} & \multicolumn{1}{c|}{\textbf{96.4}} & \multicolumn{1}{c|}{96.9} & \multicolumn{1}{c}{96.1}  \\ \bottomrule
\end{tabular}
\caption{The performance of zero-shot Chinese cross-modal retrieval with different architecture}
\label{details_results}
\end{table}

\end{document}